# Identifiability of Causal Graphs using Functional Models

**Jonas Peters**\*
Max Planck Institute
for Intelligent Systems,
Tübingen, Germany

**Joris M. Mooij**†
Institute for Computing and
Information Sciences,
Radboud University Nijmegen,
The Netherlands

**Dominik Janzing**\*
Max Planck Institute
for Intelligent Systems,
Tübingen, Germany

**Bernhard Schölkopf**\*
Max Planck Institute
for Intelligent Systems,
Tübingen, Germany

## Abstract

This work addresses the following question: Under what assumptions on the data generating process can one infer the causal graph from the joint distribution? The approach taken by conditional independence-based causal discovery methods is based on two assumptions: the *Markov condition* and *faithfulness*. It has been shown that under these assumptions the causal graph can be identified up to Markov equivalence (some arrows remain undirected) using methods like the PC algorithm. In this work we propose an alternative by defining *Identifiable Functional Model Classes* (IFMOCs). As our main theorem we prove that if the data generating process belongs to an IFMOC, one can identify the complete causal graph. To the best of our knowledge this is the first identifiability result of this kind that is not limited to linear functional relationships. We discuss how the IFMOC assumption and the Markov and faithfulness assumptions relate to each other and explain why we believe that the IFMOC assumption can be tested more easily on given data. We further provide a practical algorithm that recovers the causal graph from finitely many data; experiments on simulated data support the theoretical findings.

## 1 Introduction

Inferring the causal structure of a set of random variables given a sample from the joint distribution is an important problem in science.

*Conditional independence-based* methods like the PC algorithm [Spirtes et al., 2001] are based on the following idea: Assume that the joint distribution is Markov with respect to the causal graph, that means each variable is independent of all its non-descendants given its parents. If two variables, for example, are always dependent, no matter what other variables one conditions on, these two variables must be adjacent. Thus, properties of the joint distribution can help to infer parts of the graph structure. If one additionally assumes faithfulness (that means *all* conditional independences in the joint distribution are entailed by the Markov condition), one can use further reasonings like: If two variables are independent there is no collider-free path between them. Obviously, many more rules like this can be exploited. It is clear, however, that these ideas are not able to distinguish between two graphs that entail exactly the same set of (conditional) independences, i.e. between Markov equivalent graphs. Since many Markov equivalence classes contain more than one graph, conditional independence-based methods leave some arrows undirected and cannot uniquely identify the true causal graph.

Consider the case of two dependent random variables. Conditional independence-based methods cannot recover the graph since there is no (conditional) independence statement; $X \to Y$ and $Y \to X$ are Markov equivalent. Recently, Shimizu et al. [2006], Hoyer et al. [2009], Peters et al. [2010] suggest the following procedure to tackle this particularly challenging problem: Whenever the joint distribution $\mathbb{P}^{(X,Y)}$ allows an *additive noise model (ANM)* in one direction, i.e., there is a function $f$ and a noise variable $N$, such that[1]

$$Y = f(X) + N, \quad N \perp\!\!\!\perp X,$$

but not in the other, one infers the former direction to be the causal one (here: $X \to Y$). They further show that under mild conditions (essentially some combinations of $f$, $\mathbb{P}^X$ and $\mathbb{P}^N$ have to be excluded) the model is identifiable. This means that whenever there is an ANM from $X$ to $Y$ the joint distribution does not allow for an ANM from $Y$ to $X$. In this work we call these cases "bivariate identifiable". Another example of a bivariate identifiable model class are post-nonlinear models [Zhang and Hyvärinen, 2009].

---

\*{jpeters, janzing, bs}@tuebingen.mpg.de
†j.mooij@cs.ru.nl

[1] $N \perp\!\!\!\perp X$ means $N$ and $X$ are statistical independent.

Based on bivariate identifiability we define *Identifiable Functional Model Classes (IFMOCs)*, which we use to model distributions of more than two random variables. As a main result of this paper we prove that whenever a data generating process belongs to an IFMOC, one can recover the true causal graph from the joint distribution. To the best of our knowledge this is the first identifiability statement of this kind that allows for nonlinear interactions.

Analogously to the two-variable case described above, practical methods for causal inference using ANMs have been suggested for the multivariate case [Hoyer et al., 2009, Zhang and Hyvärinen, 2009, Mooij et al., 2009, Tillman et al., 2009]: whenever a graph models the data the method infer this structure as the causal graph. Our results fill a theoretical gap that has remained open so far: except for the linear case [Shimizu et al., 2006] the corresponding identifiability problem had not been solved yet.

What happens if the IFMOC assumption is not satisfied? Janzing and Steudel [2010] argue that the true data generating process should not fit an ANM *only* in the wrong causal direction. We believe that there is an analogous statement for the multivariate case, i.e., we do not expect the process to belong to an IFMOC *only* for a different ordering of the variables. If one accepts this belief one can test whether the IFMOC assumption is valid: if it is not, one can either fit none or multiple models to the data. In order to exploit this deliberation, we provide an algorithm that outputs *all* structures that fit the data. In contrast, faithfulness in its general form cannot be tested on given data, although Zhang and Spirtes [2007] decompose faithfulness into parts, some of which can be tested.

The paper is organized as follows: Section 2 provides the two identifiability results and discusses their assumptions. Section 3 proves of our main theorem. Section 4 provides an algorithm that identifies the causal graph given a data set and Section 5 contains experiments on artificial data.

## 2 Identifiability Results

Let $(X_i)_{i \in \mathbf{V}}$ be a finite family of random variables. This work addresses the following problem:

**Problem** *Given i.i.d. samples from a joint distribution $\mathbb{P}^{(X_i), i \in \mathbf{V}}$, infer the true causal DAG $\mathcal{G}_c$ of the process that generated the data.*

The *causal graph* $\mathcal{G}_c$ is constructed by drawing arrows from nodes to their direct effects. Throughout the paper we assume an acyclic $\mathcal{G}_c$, causal sufficiency (the absence of latent common causes) and no selection bias (conditioning on common effects). Specifically, our goal is to investigate under what assumptions this problem is solvable. Section 2.2 describes how conditional independence-based methods attempt to tackle this problem. In Section 2.3 we present the alternative we propose and Section 2.4 compares the assumptions. But first, we introduce concepts of graphical models that will be used throughout the paper.

### 2.1 Directed Acyclic Graphs

Let $(X_i)_{i \in \mathbf{V}}$ be a finite family of random variables and $\mathcal{G} = (\mathbf{V}, \mathcal{E})$ be a directed acyclic graph (DAG), with edges $\mathcal{E} \subseteq \mathbf{V}^2$. In a slight abuse of notation we often speak of both $X_i$ and $i$ being the nodes (or vertices) of $\mathcal{G}$. The following definitions are well-known and can be found in [Spirtes et al., 2001, Koller and Friedman, 2009], for example.

**Definition 1**
- $X_i$ is called a parent of $X_j$ if $(i, j) \in \mathcal{E}$ and a child if $(j, i) \in \mathcal{E}$. The set of parents of $X_j$ is denoted $\mathbf{PA}_j^{\mathcal{G}}$, the set of its children by $\mathbf{CH}_j^{\mathcal{G}}$.
- *Three nodes are called an* immorality *if one node is a child of the two others, which themselves are not adjacent. If $(i, j) \in \mathcal{E}$ and $(k, j) \in \mathcal{E}$, $j$ is called a* collider.
- *A* path *in graph $\mathcal{G}$ is a sequence of (at least two) distinct vertices $X_{i_1}, \ldots, X_{i_n}$, such that $(i_k, i_{k+1}) \in \mathcal{E}$ or $(i_{k+1}, i_k) \in \mathcal{E}$ for all $k = 1, \ldots, n-1$. If for all $k$ the former holds we speak of a* directed path *between $X_{i_1}$ and $X_{i_n}$ and call $X_{i_n}$ a descendant of $X_{i_1}$. We denote all descendants of $X_i$ by $\mathbf{DE}_i^{\mathcal{G}}$ and all non-descendants of $X_i$ by $\mathbf{ND}_i^{\mathcal{G}}$.*
- *A path between $X_{i_1}$ and $X_{i_n}$ is blocked by a set $\mathbf{S}$ (with neither $X_{i_1}$ nor $X_{i_n}$ in this set) whenever there is a node $X_{i_k}$, such that one of the following two possibilities hold:*
  1. $X_{i_k} \in \mathbf{S}$ and
  $$X_{i_{k-1}} \to X_{i_k} \to X_{i_{k+1}} \text{ or}$$
  $$X_{i_{k-1}} \leftarrow X_{i_k} \leftarrow X_{i_{k+1}} \text{ or}$$
  $$X_{i_{k-1}} \leftarrow X_{i_k} \to X_{i_{k+1}}$$
  2. $X_{i_{k-1}} \to X_{i_k} \leftarrow X_{i_{k+1}}$ *and neither $X_{i_k}$ nor any of its descendants is in $\mathbf{S}$.*

*We say that two disjoint subsets of vertices $\mathbf{A}$ and $\mathbf{B}$ are* d-separated *by a third (also disjoint) subset $\mathbf{S}$ if every path between nodes in $\mathbf{A}$ and $\mathbf{B}$ is blocked by $\mathbf{S}$.*

- *The joint distribution $\mathbb{P}^{(X_i), i \in \mathbf{V}}$ is said to be* Markov *with respect to the DAG $\mathcal{G}$ if*

  $$\mathbf{A}, \mathbf{B} \text{ d-sep. by } \mathbf{C} \Rightarrow \mathbf{A} \perp\!\!\!\perp \mathbf{B} \mid \mathbf{C}$$

  *for all disjoint sets $\mathbf{A}, \mathbf{B}, \mathbf{C}$.*
- $\mathbb{P}^{(X_i), i \in \mathbf{V}}$ *is said to be* faithful *to the DAG $\mathcal{G}$ if*

  $$\mathbf{A}, \mathbf{B} \text{ d-sep. by } \mathbf{C} \Leftarrow \mathbf{A} \perp\!\!\!\perp \mathbf{B} \mid \mathbf{C}$$

  *for all disjoint sets $\mathbf{A}, \mathbf{B}, \mathbf{C}$.*
- *We call two graphs* Markov equivalent *if they satisfy the same set of d-separations, that means the Markov condition entails the same set of (conditional) independence conditions.*

**Notation:** In the remainder of this article we use the following notation: $p_X(x)$ denotes the pdf (or pmf) of a random variable $X$, $p_\mathbf{S}(x_\mathbf{S})$ denotes the joint pdf (or pmf) for a set of random variables $\mathbf{S}$ evaluated at the point $x_\mathbf{S}$. We will assume that $\mathbb{P}^{(X_i), i \in \mathbf{V}}$ is absolutely continuous with respect to either the Lebesgue measure or the counting measure (i.e., either we have a pdf or a pmf). Then $Y|_{X=x}$ is a RV that corresponds to the conditional density $p_{Y \mid X=x}(y) = \frac{p_{X,Y}(x,y)}{p_X(x)}$.

## 2.2 Identifiability for Conditional Independence-Based Methods

All conditional independence-based (CIB) methods we are aware of make the following

**Assumption 1 (Markov and Faithfulness)**
*Assume $\mathbb{P}^{(X_i), i \in \mathbf{V}}$ is Markov and faithful with respect to the true causal DAG $\mathcal{G}_c$.*

In Section 2.4 we discuss Assumption 1 in more detail. Given that one observes i.i.d. data sampled from $\mathbb{P}^{(X_i), i \in \mathbf{V}}$, Pearl [2009], Spirtes et al. [2001], Meek [1995] show how one can exploit conditional independences in the data for (partially) reconstructing the graph $\mathcal{G}_c$, e.g., by using the PC algorithm. The graph can only be recovered up to Markov equivalence classes.

**Theorem 1 (Correctness of CIB Methods)**
*Under Assumption 1 one can identify the Markov equivalence class of the true causal DAG from the joint distribution.*

Note that the first step of the PC algorithm determines the variables that are adjacent. Therefore one has to test whether two variables are dependent given *any* other subset of variables, which results in conditional independence tests with conditioning sets of up to $\#\mathbf{V} - 2$ variables. Such tests are difficult to perform in practice [e.g. Bergsma, 2004].

Summarizing, from our perspective this approach has the following drawbacks: (1) We can identify the true causal DAG only up to Markov equivalence classes. (2) Conditional independence testing, especially with a large conditioning set, is difficult in practice. (3) The faithfulness condition in its general form cannot be tested given the data. (4) If faithfulness is violated we do not have any guarantees that the inferred graph(s) will be close to the original.

## 2.3 Identifiability using Functional Models

First, we define *functional models* [e.g. Chapter 1.4 Pearl, 2009]) that are also known as *Structural Equation Models*:

**Definition 2 ($\mathcal{F}$-FMOC)**
- $\#\mathbf{V}$ *equations*

  $$X_i = f_i(\mathbf{PA}_i, N_i) \qquad i \in \mathbf{V}$$

  *with sets of variables $\mathbf{PA}_i \subseteq \mathbf{V} \setminus \{X_i\}$ and noise distributions $\mathbb{P}^{N_i}$ are called a* functional model *if the $(N_i)_{i \in \mathbf{V}}$ are jointly independent and the graph that is obtained by drawing arrows from all elements of $\mathbf{PA}_i$ to $X_i$ (for each $i \in \mathbf{V}$) is acyclic.*
- *Given a set of functions*

  $$\mathcal{F} \subset \{f \mid f : \mathbb{R}^m \to \mathbb{R} \text{ for any } 2 \leq m \leq \#\mathbf{V}\}$$

  *we call a set of functional models a* functional model class with function class $\mathcal{F}$ ($\mathcal{F}$-FMOC) *if each of the functional models satisfies $f_i \in \mathcal{F}$ for all $i \in \mathbf{V}$ and induces $\mathbb{P}^{(X_i), i \in \mathbf{V}}$ that is absolutely continuous with respect to the Lebesgue measure or the counting measure.*

Note that each functional model induces a unique joint distribution $\mathbb{P}^{(X_i), i \in \mathbf{V}}$. Some recent causal discovery methods distinguish between cause and effect by means of the following observation. For some classes of bivariate functional models it has been shown that the structure of the model is in the "generic case" identifiable from the joint distribution: Consider, for example, only linear and additive functions $f(x,n) = a \cdot x + n$ and non-Gaussian noise. Then Shimizu et al. [2006] show that if $Y = f(X, N_Y)$ holds with $N_Y \perp\!\!\!\perp X$, one cannot find any function $g$ such that $X = g(Y, N_X)$ with $N_X \perp\!\!\!\perp Y$. Thus, we will call the set of all triples $(f, \mathbb{P}^X, \mathbb{P}^N)$ of linear functions and non-Gaussian distributions *bivariate identifiable*. Hoyer et al. [2009], Peters et al. [2010] show a similar result for non-linear additive functions $f(x,n) = g(x) + n$, and Zhang and Hyvärinen [2009] for post-nonlinear models $f(x,n) = h(g(x) + n)$ with invertible $h$. Writing $\mathcal{F}_{|_2} := \{f \in \mathcal{F} \mid f : \mathbb{R}^2 \to \mathbb{R}\}$ we can now generalize these ideas and define:

**Definition 3 (Bivariate Identifiable Set)** Let $\mathcal{F}$ be a set of functions as above. We call a set $\mathcal{B} \subseteq \mathcal{F}_{|2} \times \mathcal{P}_\mathbb{R} \times \mathcal{P}_\mathbb{R}$ containing combinations of functions $f \in \mathcal{F}_{|2}$ and distributions $\mathbb{P}^X$, $\mathbb{P}^{N_Y}$ of input $X$ and noise $N_Y$ bivariate identifiable in $\mathcal{F}$ if

$$(f, \mathbb{P}^X, \mathbb{P}^{N_Y}) \in \mathcal{B} \text{ and } Y = f(X, N_Y), N_Y \perp\!\!\!\perp X$$
$$\Rightarrow \quad \nexists g \in \mathcal{F}_{|2} : \quad X = g(Y, N_X), N_X \perp\!\!\!\perp Y$$

holds. Additionally we require

$$f(X, N_Y) \not\perp\!\!\!\perp X \qquad (1)$$

for all $(f, \mathbb{P}^X, \mathbb{P}^{N_Y}) \in \mathcal{B}$ with $N_Y \perp\!\!\!\perp X$.

The first part of the definition requires that we cannot simultaneously fit both directions (think of $\mathcal{F}$ being the class of linear ANMs and $\mathcal{B}$ being all of those models, where input and noise are *not* jointly Gaussian). The left hand side of (1) corresponds to the effect, the right hand side to the cause. In the bivariate case one can imagine that we do not want them to be independent. Section 2.4 discusses this assumption.

Note further that the function class needs to be restricted for the definition to be non-trivial, because for any joint distribution of $(X, Y)$ we can find a function $f$ and a noise $N_Y \perp\!\!\!\perp X$, such that $Y = f(X, N_Y)$ [Darmois, 1951]. Proving that a set is bivariate identifiable is not trivial. The following lemma presents identifiability results that have been reported in literature. In order to improve readability, we describe the classes and mention only the most important counterexamples. We denote all other exceptions by the sets $\tilde{B}_i$, which mostly contain constant functions and other, "non-generic" cases.

**Lemma 1** *The following sets have been shown to be bivariate identifiable ($\tilde{m} \in \mathbb{N}$):*
*(i) linear ANMs: $\mathcal{F}_1 = \{f(x, n) = ax + n\}$*

$$\mathcal{B}_1 = \{(X, N) \text{ not both Gaussian}\} \setminus \tilde{B}_1$$

*(ii) discrete ANMs: $\mathcal{F}_2 = \{f(x, n) \equiv \phi(x) + n \bmod \tilde{m}\}$*

$$\mathcal{B}_2 = \{(\phi, X) \text{ not affine and uniform}\} \setminus \tilde{B}_2$$

*(iii) nonlinear ANMs: $\mathcal{F}_3 = \{f(x, n) = \phi(x) + n\}$*

$$\mathcal{B}_3 = \{(\phi, X, N) \text{ not lin., Gauss, Gauss}\} \setminus \tilde{B}_3$$

*(iv) post-nonlin.: $\mathcal{F}_4 = \{f(x, n) = \psi(\phi(x) + n), \psi \text{ inv.}\}$*

$$\mathcal{B}_4 = \{(\psi, \phi, N) \text{ not lin., lin., Gauss}\} \setminus \tilde{B}_4$$

**Proof.** Shimizu et al. [2006], Peters et al. [2010], Hoyer et al. [2009] and Zhang and Hyvärinen [2009] provide proofs and the precise definitions of the sets $\tilde{B}_i$ for (i)-(iv), respectively. □

In our final definition we generalize the concept of bivariate identifiable to more than two variables:

**Definition 4 (($\mathcal{B}, \mathcal{F}$)-IFMOC)** *Let $\mathcal{B}$ be bivariate identifiable in $\mathcal{F}$. We call an $\mathcal{F}$-FMOC a $(\mathcal{B}, \mathcal{F})$-Identifiable Functional Model Class, for short $(\mathcal{B}, \mathcal{F})$-IFMOC, if for all its functional models*

$$X_i = f_i(\mathbf{PA}_i, N_i), \qquad i \in \mathbf{V}$$

*for all $i \in \mathbf{V}$, $j \in \mathbf{PA}_i$ and for all $x_{\mathbf{PA}_i \setminus \{j\}}$, we have*

$$f_i(x_{\mathbf{PA}_i \setminus \{j\}}, \underbrace{\cdot}_{X_j}, \underbrace{\cdot}_{N_i}) \in \mathcal{F}_{|2} . \qquad (2)$$

*Additionally, for all sets $\mathbf{S} \subseteq \mathbf{V}$ with $\mathbf{PA}_i \setminus \{j\} \subseteq \mathbf{S} \subseteq \mathbf{ND}_i \setminus \{i, j\}$, there exists an $x_\mathbf{S}$ with $p_\mathbf{S}(x_\mathbf{S}) > 0$ and*

$$\left( f_i(x_{\mathbf{PA}_i \setminus \{j\}}, \underbrace{\cdot}_{X_j}, \underbrace{\cdot}_{N_i}), \mathbb{P}^{X_j \mid X_\mathbf{S} = x_\mathbf{S}}, \mathbb{P}^{N_i} \right) \in \mathcal{B} . \quad (3)$$

Thus, an $(\mathcal{B}, \mathcal{F})$-IFMOC consists of many functional models, which are defined in Defintion 2.

**Example 1**
- *In the bivariate case ($\#V = 2$), in Definition 4 we have $\mathbf{S} = \emptyset$ and thus equation (2) is always satisfied. (3) then reads that the triple $(f_2, \mathbb{P}^{X_1}, \mathbb{P}^{N_2})$ is in the bivariate identifiable set $\mathcal{B}$ (if $\mathbf{PA}_2 = \{X_1\}$).*
- *For more than two variables one can exploit Lemma 1. For ANMs equation (2) holds: the functions remain additive in the noise if some arguments are fixed. If one further uses linear ANMs $\mathcal{F} = \mathcal{F}_1$, for example, and restricts $\mathcal{B}$ to contain only non-Gaussian noise, also (3) holds and we recover LiNGAM [Shimizu et al., 2006]. Using the other $\mathcal{F} \neq \mathcal{F}_1$ from Lemma 1 we obtain analogous results for the nonlinear case.*

Now we are able to state our main theoretical result:

**Theorem 2** *Assume that $\mathbb{P}^{(X_i), i \in \mathbf{V}}$ is induced by a functional model from a $(\mathcal{B}, \mathcal{F})$-IFMOC with graph $\mathcal{G}$. Then it cannot be induced by a functional model from the same $(\mathcal{B}, \mathcal{F})$-IFMOC that corresponds to a different graph $\mathcal{G}' \neq \mathcal{G}$.*

The proof can be found in Section 3. In the context of causal inference we have the following reformulation:

**Assumption 2 (IFMOC Assumption)** *Assume that the data generating mechanism belongs to an $(\mathcal{B}, \mathcal{F})$-IFMOC with graph $\mathcal{G} = \mathcal{G}_c$ (i.e., $\mathbf{PA}_i^\mathcal{G}$ are the direct causes of $X_i$).*

**Corollary 1** *Under Assumption 2 we can identify the true causal DAG $\mathcal{G}_c$ from the joint distribution $\mathbb{P}^{(X_i), i \in \mathbf{V}}$.*

We do not claim that each natural process satisfies Assumption 2, only that *if it does*, we can *then* recover the true causal relationships from the joint distribution. In our opinion, this approach provides the following advantages: (1) We can identify the true causal graph even within the Markov equivalence class. (2) One can use IFMOCs to identify non-faithful causal models (even those "undetectable" versions of unfaithfulness mentioned in Section 2.4), for which conditional independence-based methods usually fail. (3) In our opinion the IFMOC assumption can be tested given the data (see Section 2.4). (4) A functional model contains more information than the corresponding causal Markov DAG: some counterfactual statements, for example, can only be deduced from the functional model and not from the causal Markov DAG [e.g. 1.4.4 in Pearl, 2009].
Note that our result already includes discrete models, but only works for non-deterministic data.

### 2.4 Discussion of the Assumptions

We briefly discuss the differences and similarities between Assumptions 1 and 2.

**Markov condition.** Assume a functional model for $X_1, \ldots, X_n$. Then Pearl [2009] shows in Theorem 1.4.1 that the joint distribution is Markov with respect to the corresponding graph. Therefore, this part of the assumptions is common to conditional independence-based approaches and IFMOC based approaches.

**Faithfulness.** Zhang and Spirtes [2007] analyse the testability of faithfulness. They decompose faithfulness into adjacency-faithfulness (two adjacent variables are dependent conditional on any set of other variables) and orientation-faithfulness (a structure $X \to Y \leftarrow Z$ renders $X$ and $Z$ dependent given any set that contains $Y$ and a structure $X - Y - Z$ with arrows other than above renders $X$ and $Z$ dependent given any set of variables that *does not* contain $Y$). They prove, for example, that under the assumption of Markov condition and adjacency faithfulness, any violation of orientation-faithfulness is detectable. However, some violations of adjacency faithfulness (e.g. $X \to Y \to Z$ and $X \to Z$ with $X$ and $Z$ independent) cannot be detected because they are faithful to an alternative structure ($X \to Y \leftarrow Z$).

**IFMOC assumption.** The assumptions made by our approach can be violated in different ways. (1) The true data generating process belongs to an FMOC (e.g., linear interactions and additive noise), but not to an IFMOC (e.g., the interactions are linear and all variables are Gaussian distributed). In this case the joint distribution allows several representations that lead to different causal graphs. Thus, the method could output: "More than one graph possible, no answer proposed.". However, if we are willing to assume faithfulness, we can recover the Markov equivalence class by choosing the DAGs with the minimal number of edges and thus obtain asymptotically the same results as the PC algorithm.[2] (2) The true process does not belong to an FMOC. Here, the method would not be able to fit the data. Therefore the method should output: "Bad model fit. Try a different model class." (3) The true process does not belong to an FMOC, but belongs to an IFMOC with a different graph than the true causal graph (e.g., $X \to Y$ is the ground truth and the joint distribution does not allow an ANM from $X$ to $Y$, but only from $Y$ to $X$). This is the only situation, in which our method fails and gives a wrong answer. We would like to argue, however, that this case is unlikely in the following sense: Janzing and Steudel [2010] use the concept of Kolmogorov complexity to show that it can only happen if the cause distribution $p(\text{cause})$ and the conditional distribution of the effect given the cause $p(\text{effect} \,|\, \text{cause})$ are matched in a precise way, whereas one rather expects input and mechanism to be most often "independent" [Lemeire and Dirkx, 2006, Janzing and Schölkopf, 2010]. Janzing and Steudel [2010] only consider the bivariate case, but we expect a similar statement to hold in general.

**Faithfulness vs. IFMOC assumption.** There is a connection between equation (1) and faithfulness. In the context of an IFMOC, (1) basically reads as Lemma 4 (see below). If the latter is violated, this is also a violation of faithfulness (in this sense, faithfulness is stronger). We even show in Proposition 2 that Lemma 4 implies *causal minimality*, a weak form of faithfulness [Spirtes et al., 2001]. Causal minimality states that a joint distribution is not Markov with respect to a strict subgraph of the true causal graph $\mathcal{G}_c$. Further, if $g(x, n) = n$ lies in $\mathcal{F}_{|_2}$, (1) is satisfied: If $Y = f(X, N_Y) \perp\!\!\!\perp X$ were true, $X = g(Y, N_X)$ with $N_X = X$ would be a valid backward model.

## 3 Proof

We proceed with the proof of Theorem 2. The lemmata are proved in the appendix.

**Lemma 2** *Let $Y, Z, N, S$ be random variables with continuous joint density $p_{Y,Z,N,S}(y, z, n, s)$ with respect to some product measure (all random variables here can be multivariate). Let $f : \mathcal{Y} \times \mathcal{Z} \times \mathcal{N} \to \mathcal{X}$ be a measurable function. If $N \perp\!\!\!\perp (Y, Z, S)$ then for all $z \in \mathcal{Z}, s \in S$ with $p_{Z,S}(z, s) > 0$:*

$$f(Y, Z, N)_{\,|\, Z=z, S=s} = f(Y_{\,|\, Z=z, S=s}, z, N)$$

---

[2]Proposition 1 in the appendix proves this statement.

**Lemma 3** *Consider a functional model with corresponding DAG $\mathcal{G}$ and a random variable $X$. If $\mathbf{S} \subseteq \mathbf{ND}_X^{\mathcal{G}}$ then $N_X \perp\!\!\!\perp \mathbf{S}$.*

**Lemma 4** *Consider an instance of an IFMOC with DAG $\mathcal{G}$, a variable $B$ and one of its parents $A$. For all sets $\mathbf{S}$ with $\mathbf{PA}_B^{\mathcal{G}} \setminus \{A\} \subseteq \mathbf{S} \subseteq \mathbf{ND}_B^{\mathcal{G}}$ we have*

$$B \not\perp\!\!\!\perp A \mid \mathbf{S}$$

**Proof of Theorem 2.** We assume that there are two instances of an IFMOC that both induce $\mathbb{P}^{(X_i), i \in \mathbf{V}}$, one with graph $\mathcal{G}$, the other with graph $\mathcal{G}'$. We will show that $\mathcal{G} = \mathcal{G}'$. Since DAGs do not contain any cycles, we always find nodes that have no descendants (start a directed path at some node: after at most $\#\mathbf{V} - 1$ steps you reach a node without a child). Eliminating such a node from the graph leads to a DAG, again; we can discard further nodes without children in the new graph. We repeat this process for all nodes that have no children in both $\mathcal{G}$ and $\mathcal{G}'$ and have the same parents in both graphs. If we end up with no nodes left, the two graphs are identical and we are done. Otherwise, we end up with two smaller graphs that we again call $\mathcal{G}$ and $\mathcal{G}'$ and a node $X$ that has no children in $\mathcal{G}$ and either $\mathbf{PA}_X^{\mathcal{G}} \neq \mathbf{PA}_X^{\mathcal{G}'}$ or $\mathbf{CH}_X^{\mathcal{G}'} \neq \emptyset$. We will show that this leads to a contradiction. Importantly, because of the Markov property of $\mathcal{G}$, all other nodes are independent of $X$ given $\mathbf{PA}_X^{\mathcal{G}}$:

$$X \perp\!\!\!\perp \mathbf{V} \setminus (\mathbf{PA}_X^{\mathcal{G}} \cup \{X\}) \mid \mathbf{PA}_X^{\mathcal{G}} \qquad (4)$$

To make the arguments easier to understand, we introduce the following notation (see also Figure 1): We partition $\mathcal{G}$-parents of $X$ into $\mathbf{Y}, \mathbf{Z}$ and $\mathbf{W}$. Here, $\mathbf{Z}$ are also $\mathcal{G}'$-parents of $X$, $\mathbf{Y}$ are $\mathcal{G}'$-children of $X$ and $\mathbf{W}$ are not adjacent to $X$ in $\mathcal{G}'$. We denote with $\mathbf{D}$ the $\mathcal{G}'$-parents of $X$ that are not adjacent to $X$ in $\mathcal{G}$ and by $\mathbf{E}$ the $\mathcal{G}'$-children of $X$ that are not adjacent to $X$ in $\mathcal{G}$. Thus: $\mathbf{PA}_X^{\mathcal{G}} = \mathbf{Y} \cup \mathbf{Z} \cup \mathbf{W}$, $\mathbf{CH}_X^{\mathcal{G}} = \emptyset$, $\mathbf{PA}_X^{\mathcal{G}'} = \mathbf{Z} \cup \mathbf{D}$, $\mathbf{CH}_X^{\mathcal{G}'} = \mathbf{Y} \cup \mathbf{E}$.

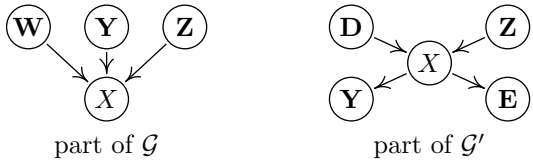

Figure 1: Nodes adjacent to $X$ in $\mathcal{G}$ and $\mathcal{G}'$.

Consider $\mathbf{T} := \mathbf{W} \cup \mathbf{Y}$. We distinguish two cases:

(i) $\mathbf{T} = \emptyset$. Then there must be a node $D \in \mathbf{D}$ or a node $E \in \mathbf{E}$, otherwise $X$ would have been discarded.

1. If there is a $D \in \mathbf{D}$ then (4) implies $X \perp\!\!\!\perp D \mid \mathbf{S}$ for $\mathbf{S} := \mathbf{Z} \cup \mathbf{D} \setminus \{D\}$, which contradicts Lemma 4 (applied to $\mathcal{G}'$).

2. If $\mathbf{D} = \emptyset$ and there is $E \in \mathbf{E}$ then $E \perp\!\!\!\perp X \mid \mathbf{S}$ holds for $\mathbf{S} := \mathbf{Z} \cup \mathbf{PA}_E^{\mathcal{G}'} \setminus \{X\}$, which also contradicts Lemma 4 (note that $\mathbf{Z} \subseteq \mathbf{ND}_E^{\mathcal{G}'}$ to avoid cycles).

(ii) $\mathbf{T} \neq \emptyset$. Then $\mathbf{T}$ contains a "$\mathcal{G}'$-youngest" node with the property that there is no directed $\mathcal{G}'$-path from this node to any other node in $\mathbf{T}$. This node may not be unique.

1. Suppose that some $W \in \mathbf{W}$ is such a youngest node. Consider the DAG $\tilde{\mathcal{G}}'$ that equals $\mathcal{G}'$ with additional edges $Y \to W$ and $W' \to W$ for all $Y \in \mathbf{Y}$ and $W' \in \mathbf{W} \setminus \{W\}$. In $\tilde{\mathcal{G}}'$ $X$ and $W$ are not adjacent. Thus we find a set $\tilde{\mathbf{S}}$ such that $\tilde{\mathbf{S}}$ $d$-separates $X$ and $W$ in $\tilde{\mathcal{G}}'$; indeed, one can take $\tilde{\mathbf{S}} := \left(\mathbf{CH}_X^{\tilde{\mathcal{G}}'} \cup \mathbf{PA}^{\tilde{\mathcal{G}}'}(\mathbf{CH}_X^{\tilde{\mathcal{G}}'})\right) \setminus \left(\mathbf{U} \cup \mathbf{DE}^{\tilde{\mathcal{G}}'}(\mathbf{U})\right)$ with $\mathbf{U} = \mathbf{CH}_X^{\tilde{\mathcal{G}}'} \cap \mathbf{CH}_W^{\tilde{\mathcal{G}}'}$. Then also $\mathbf{S} = \tilde{\mathbf{S}} \cup \{\mathbf{Y}, \mathbf{Z}, \mathbf{W} \setminus \{W\}\}$ $d$-separates $X$ and $W$ in $\tilde{\mathcal{G}}'$.

   Indeed: All $Y \in \mathbf{Y}$ are already in $\tilde{\mathbf{S}}$ in order to block $X \to Y \to W$. Suppose there is a $\tilde{\mathcal{G}}'$-path that is blocked by $\tilde{\mathbf{S}}$ and unblocked if we add $Z$ and $W'$ nodes to $\tilde{\mathbf{S}}$. How can we unblock a path by including more nodes? The path $(X \cdots V_1 \cdots U_1 \cdots W$ in Figure 2) must contain a collider $V_1$ that is an ancestor of a $Z$ with $V_1, \ldots, V_m, Z \notin \tilde{\mathbf{S}}$ and corresponding nodes $U_i$ for a $W'$ node. Choose $V_1$ and $U_1$ on the given path so close to each other such that there is no such a collider in between. If there is no $V_1$, choose $U_1$ close to $X$, if there is no $U_1$, choose $V_1$ close to $W$. Now the path $X \leftarrow Z \cdots V_1 \cdots U_1 \cdots W' \to W$ is unblocked given $\tilde{\mathbf{S}}$, which is a contradiction to $\tilde{\mathbf{S}}$ $d$-separates $X$ and $W$.

   But then $\mathbf{S}$ $d$-separates $X$ and $W$ in $\mathcal{G}'$, too and we have $X \perp\!\!\!\perp W \mid \mathbf{S}$ which contradicts Lemma 4 (applied to $\mathcal{G}$).

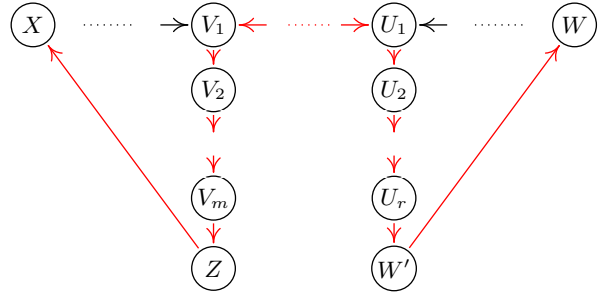

Figure 2: Assume the path $X \cdots V_1 \cdots U_1 \cdots W$ is blocked by $\tilde{\mathbf{S}}$, but unblocked if we include $Z$ and $W'$. Then the red path is unblocked given $\tilde{\mathbf{S}}$.

2. Therefore, the $\mathcal{G}'$-youngest node in $\mathbf{T}$ must be some $Y \in \mathbf{Y}$. We define $\mathbf{S} := \mathbf{PA}_X^{\mathcal{G}} \setminus \{Y\} \cup \mathbf{PA}_Y^{\mathcal{G}'} \setminus \{X\}$. Clearly, $\mathbf{S} \subseteq \mathbf{ND}_X^{\mathcal{G}}$ since $X$ does not have any descendants in $\mathcal{G}$. Further, $\mathbf{S} \subseteq \mathbf{ND}_Y^{\mathcal{G}'}$ because $Y$ is the youngest under all $W \in \mathbf{W}$ and $Y \in \mathbf{Y} \setminus \{Y\}$ by construction and any directed path from $Y$ to $Z \in \mathbf{Z}$ would introduce a cycle in $\mathcal{G}'$. Ergo, $\{Y\} \cup \mathbf{S} \subseteq \mathbf{ND}_X^{\mathcal{G}}$ and $\{X\} \cup \mathbf{S} \subseteq \mathbf{ND}_Y^{\mathcal{G}'}$. Lemma 3 gives us $N_X \perp\!\!\!\perp (Y, \mathbf{S})$ and $N_Y \perp\!\!\!\perp (X, \mathbf{S})$ and we can thus apply Lemma 2. From $\mathcal{G}$ we find

$$X_{\mid X_{\mathbf{S}} = x_{\mathbf{S}}} = f_X(x_{\mathbf{PA}_X^{\mathcal{G}} \setminus \{Y\}}, Y_{\mid X_{\mathbf{S}} = x_{\mathbf{S}}}, N_X),$$
$$N_X \perp\!\!\!\perp Y_{\mid X_{\mathbf{S}} = x_{\mathbf{S}}}$$

and from $\mathcal{G}'$ we have

$$Y_{\mid X_{\mathbf{S}} = x_{\mathbf{S}}} = g_Y(x_{\mathbf{PA}_Y^{\mathcal{G}'} \setminus \{X\}}, X_{\mid X_{\mathbf{S}} = x_{\mathbf{S}}}, N_Y),$$
$$N_Y \perp\!\!\!\perp X_{\mid X_{\mathbf{S}} = x_{\mathbf{S}}}$$

This leads to a contradiction since according to Definition 4 we can choose $x_{\mathbf{S}}$ such that $(f_X(x_{\mathbf{PA}_X^{\mathcal{G}} \setminus \{Y\}}, \cdot, \cdot), \mathbb{P}^{Y \mid X_{\mathbf{S}} = x_{\mathbf{S}}}, \mathbb{P}^{N_X}) \in \mathcal{B}$, and $g_Y(x_{\mathbf{PA}_Y^{\mathcal{G}'} \setminus \{X\}}, \cdot, \cdot) \in \mathcal{F}$.

$\square$

## 4 Algorithm

Given a data set the main idea of the method is as follows: for each graph structure it fits the corresponding functional model from the $\mathcal{F}$-FMOC and outputs all graphs, for which the residuals are independent. If the algorithm has either no or multiple outputs, Theorem 2 proves that Assumption 2 must be violated. Algorithm 1 shows how to avoid checking all possible DAGs: it finds the sink node, disregards it and continues with the smaller graph. The algorithm is based on [Mooij et al., 2009] but outputs *all* graphs that are consistent with the data by using depth-first search: whenever there is more than one way to proceed in building the DAG, instead of choosing the one that leads to the highest p-value of the independence test (see Algorithm 1, line 8 in Mooij et al. [2009]) we keep track of all possibilities. Note that $\sigma_1, \ldots, \sigma_d$ give the causal order; they also depend on *currentcase* (omitted to improve readability). To increase robustness, we test for joint independence of the residuals at the end (not shown). The algorithm runs with any independence test and any regression method, our choices are described below. Code can be found on the homepage of the first author or of the MPI causality group.

## 5 Experiments

For regression we either use linear regression (IFMOC$_{\text{lin}}$) or Gaussian Processes as in [Hoyer et al., 2009] (IFMOC$_{\text{GP}}$). To check whether the residuals are independent of the regressors we use HSIC [Gretton et al., 2008]. For the PC algorithm we used an implementation by Tillman et al. [2009] and as a test either partial correlation (PC$_{\text{corr}}$) or "conditional HSIC" (PC$_{\text{HSIC}}$) proposed by Fukumizu et al. [2008] with 500 bootstrap samples to generate the null distribution. Ignoring problems of multiple testing we always set the significance level of statistical tests to 5%.

**Algorithm 1** Finding all possible DAGs
1: **input** data matrix $X$ of size $N \times d$, sign. value $\alpha$
2: $totalcases \leftarrow 1, currentcase \leftarrow 1$
3: $S(1) \leftarrow \{1, \ldots, d\}, jj(1) \leftarrow d, \sigma_1 \leftarrow 0$
4: **while** $currentcase \leq totalcases$ **do**
5:   **for** $j = jj(currentcase)$ **downto** $1$ **do**
6:     **for all** $i \in S$ **do**
7:       $\hat{\epsilon}_i \leftarrow \texttt{FittedNoiseValues}(X_{S \setminus \{i\}}, X_i)$
8:       $p_i \leftarrow \texttt{TestIndependence}(X_{S \setminus \{i\}}, \hat{\epsilon}_i)$
9:     **end for**
10:     $i^* \leftarrow \arg\max p_i$
11:     **if** $p_i < \alpha$ for all $i$ **then**
12:       **break**
13:     **else if** $p_i \geq \alpha$ for several $i$ **then**
14:       **increase** $totalcases$ accordingly
15:       **store** $jj, \sigma, S$ and those $i$ (except $i^*$)
16:     **end if**
17:     $\sigma_j(currentcase) \leftarrow i^*$
18:     $S(currentcase) \leftarrow S(currentcase) \setminus \{i^*\}$
19:   **end for**
20:   $currentcase \leftarrow currentcase + 1$
21: **end while**
22: **for** $currentcase = 1$ **to** $totalcases$ **do**
23:   **for** $j = 1$ **to** $d$ **do**
24:     $i \leftarrow \sigma_j$
25:     $\mathbf{PA}_i \leftarrow \{\sigma_1, \ldots, \sigma_{j-1}\}$
26:     **for** $k = 1$ **to** $j - 1$ **do**
27:       $\hat{\epsilon}_i \leftarrow \texttt{FittedNoiseValues}(X_{\mathbf{PA}_i \setminus \{\sigma_k\}}, X_i)$
28:       **if** $\texttt{TestIndependence}(X_{\mathbf{PA}_i}, \hat{\epsilon}_i) \geq \alpha$ **then**
29:         $\mathbf{PA}_i \leftarrow \mathbf{PA}_i \setminus \{\sigma_k\}$
30:       **end if**
31:     **end for**
32:   **end for**
33: **end for**
34: **output** all different DAGs
35: If #DAGs $= 0$ or $\geq 2$, **output** "I do not know."

**Data Set 1: How often do we miss faithfulness?** For sample sizes between 100 and 500,000 we simulate 500 times data from the following model:

$$X_1 = \beta_1 N_1$$
$$X_2 = \alpha_{12} X_1 + \beta_2 N_2$$
$$X_3 = \alpha_{13} X_1 + \beta_3 N_3$$
$$X_4 = \alpha_{24} X_2 + \alpha_{34} X_3 + \beta_4 N_4$$

with $N_i \stackrel{iid}{\sim} \mathcal{N}(0,1)$. We regard the left DAG as ground truth and sample the coefficients $\alpha$ uniformly between $-5$ and $5$ and $\beta$ uniformly between $0$ and $0.5$. We expect the distribution to be non-faithful only on a subset of measure 0. Indeed, given the sampled coefficients we computed all (partial) correlations and verified that all distributions were faithful to the true causal graph. For finite sample size, however, we expect some cases, where the false hypothesis of zero partial correlation is not rejected. These type 2 errors lead to wrong conclusions about the underlying graph and Figure 3 shows how often they occur in the experiments. The number decreases slowly with the sample size, but even for a sample size of $500,000$ they happen in more than 10% of the cases. Note that they would be even more frequent if one lowers the significance threshold of the test. In our experiments, other distributions for $\alpha$ and $\beta$ lead to almost identical results (not shown).

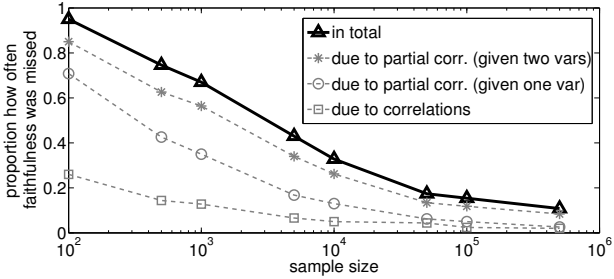

Figure 3: Data Set 1. The graph shows the proportion of cases (out of 500), where at least one (partial) correlation was falsely regarded as zero. These errors lead to wrong causal conclusions.

**Data Set 2: Both methods should work when both assumptions are met.**
We simulate 100 data sets (sample size 400) from two different structures:

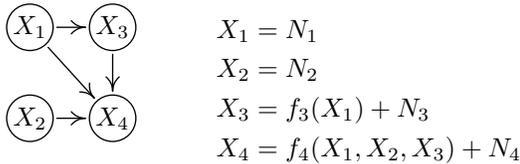

$X_1 = N_1$
$X_2 = N_2$
$X_3 = f_3(X_1) + N_3$
$X_4 = f_4(X_1, X_2, X_3) + N_4$

linear1 and nonlinear1

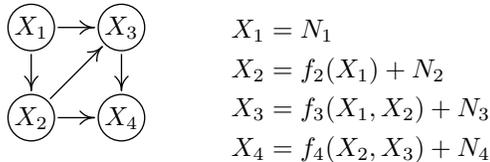

$X_1 = N_1$
$X_2 = f_2(X_1) + N_2$
$X_3 = f_3(X_1, X_2) + N_3$
$X_4 = f_4(X_2, X_3) + N_4$

linear2 and nonlinear2

with $N_i \stackrel{iid}{\sim} \mathcal{U}([-0.5, 0.5])$. We regard the drawn graphs as the true causal DAGs.
In linear1 we choose $f_i(x) = a_i^t x$ and in nonlinear1
$$f_3(x_1) = a_3 \exp(-2x_1^2) - 1$$
$$f_4(x_1, x_2, x_3) = a_{41}(x_1+1)^2 + a_{42}x_2 + a_{43}x_3.$$

For linear2 we have $f_i(x) = b_i^t x$ and for nonlinear2
$$f_2(x_1) = b_2 x_1, \ f_4(x_2, x_3) = b_{42}(x_2+1)^2 + b_{43}x_3$$
$$\text{and } f_3(x_1, x_2) = b_{31}\exp(-2x_1^2) + b_{32}x_2,$$

with $a_i, b_i \stackrel{iid}{\sim} \mathcal{U}([-2,-1] \cup [1,2])$. Table 1 shows the results. $PC_{part}$ fails for the nonlinear data sets, whereas $IMFOC_{lin}$ is undecided. The second setting is more difficult because $X_1$ and $X_4$ are only independent given $X_2$ and $X_4$ and not a single variable. Especially in this case, the proposed method seems to be more robust. Recall that for PC "correct" means having identified the Markov equivalence class containing the true graph (e.g., with an undirected arrow $X_1 - X_3$), whereas the IFMOC approach identifies the single correct DAG.

|  | lin1 | nonlin1 | lin2 | nonlin2 |
|---|---|---|---|---|
| $PC_{corr}$ | 90/10/0 | 6/94/0 | 47/53/0 | 0/100/0 |
| $PC_{HSIC}$ | 60/40/0 | 96/4/0 | 3/97/0 | 4/96/0 |
| $IFMOC_{lin}$ | 82/0/18 | 0/0/100 | 86/0/14 | 0/0/100 |
| $IFMOC_{GP}$ | 79/2/19 | 86/1/13 | 76/1/23 | 86/8/6 |

Table 1: Data Set 2. correct/wrong/undecided (out of 100). The proposed method clearly makes the least mistakes and is not always forced to take a decision.

**Data Set 3: If the distribution is not faithful, PC fails, IFMOC approach does not.**
We simulate 100 data sets (sample size 400) from

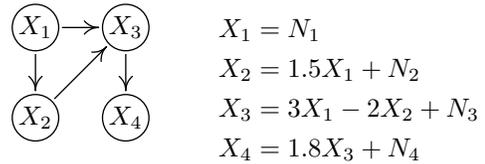

$X_1 = N_1$
$X_2 = 1.5X_1 + N_2$
$X_3 = 3X_1 - 2X_2 + N_3$
$X_4 = 1.8X_3 + N_4$

with $N_i \stackrel{iid}{\sim} \mathcal{U}([0, 0.5])$. The distribution is not faithful to the true graph (left) since $X_1 \perp\!\!\!\perp X_3$ is not entailed by the Markov condition. This is an instance of non-faithfulness that cannot be detected from the data, see Section 2.4. Out of these 100 data sets, both PC algorithms always return a wrong DAG that is not Markov equivalent to the true graph. $IFMOC_{lin}$ returns the correct DAG in 89 cases and no wrong graph.

**Data Set 4: If the data are induced by an FMOC, but not an IFMOC, both methods can return the Markov equivalence class.**
We simulate 100 data sets (sample size 400) from

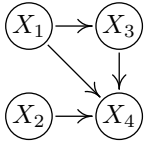

$$X_1 = 0.5N_1$$
$$X_2 = 0.5N_2$$
$$X_3 = -X_1 + 0.1N_3$$
$$X_4 = 1.5X_1 - 2X_2 + X_3 + N_4$$

with $N_i \overset{\text{iid}}{\sim} \mathcal{N}(0,1)$. The corresponding distribution is faithful to the true causal graph (left). Since the regime is Gaussian and linear, we use $\text{PC}_{\text{corr}}$ that uses partial correlation to test for conditional independence. In principle, we expect IFMOC to successfully fit functional models from different structures and to output "I do not know". If one is willing to assume faithfulness, one can output all graphs with the minimal number of edges, which correspond to the true Markov equivalence class (Section 2.4). Out of 100 data sets $\text{PC}_{\text{corr}}$ recovers the true Markov equivalence class in 47 cases (the rest is incorrect); $\text{IFMOC}_{\text{lin}}$ in 94 cases and remains undecided 6 times.

**Data Set 5: If the assumptions are violated, PC gives wrong results, IFMOC is undecided.** We simulate 100 data sets (sample size is 400) from

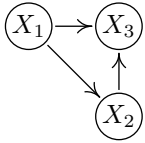

$$X_1 = N_1$$
$$X_2 = X_1 + 0.5N_2$$
$$X_3 = (X_1 - X_2) \cdot 0.5N_3$$

with $N_i \overset{\text{iid}}{\sim} \mathcal{U}([-0.5, 0.5])$. The corresponding distribution is neither faithful to the true DAG (left) nor do we expect it to satisfy an ANM. Both PC algorithms always output wrong results, whereas both IFMOC methods always output "I do not know".

## 6 Conclusion and Future Work

We proved that using identifiable functional model classes, i.e., models that are able to distinguish between $X \to Y$ and $Y \to X$, the whole true causal graph is identifiable from the joint distribution even within the Markov equivalence class. We only need to assume a weak form of faithfulness, namely causal minimality. We further built on an existing algorithm for recovering the causal graph from a finite amount of data. We tested the proposed algorithm with ANMs on simulated data sets. The experiments support our theoretical results.

Several topics remain for future work: (1) We believe that some existing methods from Bayesian Structure Learning are similar to a Bayesian version of our IFMOC approach. One may be able to apply our theoretical findings to these methods and prove their consistency. (2) In our experiments we found that the proposed method outperforms the PC algorithm even for some models that meet all assumptions; this can only occur for finitely many samples and should be investigated further. (3) It would be interesting to analyze situations, where parts of the graph satisfy the assumptions and others do not. Preliminary experiments show that some parts of the graph remain identifiable. (4) It is important to test the proposed principle for causal inference on real data sets, for which the ground truth is known, and compare it to other methods like PC.

### Acknowledgements

DJ has been supported by the DFG (SPP 1395).

**Appendix**

**Proposition 1** *Assume $\mathcal{G}_c$ is the true graph and assume $\mathbb{P}$ is faithful and Markov with respect to $\mathcal{G}_c$. If $\mathbb{P}$ is induced by a functional model from an IFMOC with $\mathcal{G}'$ as corresponding DAG we have #edges in $\mathcal{G}_c \leq$ #edges in $\mathcal{G}'$.*

**Proof.** $\mathbb{P}$ must be Markov with respect to $\mathcal{G}'$ and must thus satisfy $I_{\mathcal{G}'}$ (which we denote for all (conditional) independence equations that are induced by the graph structure of $\mathcal{G}'$). $\mathbb{P}$ must also satisfy $I_{\mathcal{G}_c}$ and since $\mathbb{P}$ is faithful wrt $\mathcal{G}_c$, we have $I_{\mathcal{G}'} \subseteq I_{\mathcal{G}_c}$. Thus: missing edges in $\mathcal{G}' \subseteq$ missing edges in $\mathcal{G}_c$ and therefore: #edges in $\mathcal{G}_c \leq$ #edges in $\mathcal{G}'$. □

**Proof of Lemma 2** First, note that the joint of $Y, Z, N, S$ satisfies:

$$p_{Y,Z,N,S}(y,z,n,s) = p_{Z,S}(z,s)p_{Y \mid Z=z, S=s}(y)p_N(n)$$

because $N \perp\!\!\!\perp (Y, Z, S)$. Consider the random variable $X := f(Y, Z, N)$. We have, for all $z \in \mathcal{Z}, s \in S$ with $p_{Z,S}(z, s) > 0$ and for all $x \in \mathcal{X}$:

$$p_{X \mid Z=z, S=s}(x) = \frac{p_{X,Z,S}(x,z,s)}{p_{Z,S}(z,s)}$$
$$= \frac{\int p_{Y,Z,N,S}(y,z,n,s)\delta(x - f(y,z,n))\,dy\,dn}{p_{Z,S}(z,s)}$$
$$= \int p_{Y \mid Z=z, S=s}(y)p_N(n)\delta(x - f(y,z,n))\,dy\,dn$$
$$= p_{f(Y_{\mid Z=z, S=s}, z, N)}(x)$$

Ergo, $X_{\mid Z=z, S=s} = f(Y_{\mid Z=z, S=s}, z, N)$ for all $z, s$ with $p_{Z,S}(z,s) > 0$. □

**Proof of Lemma 3** Write $\mathbf{S} = \{S_1, \ldots, S_k\}$. Then

$$\mathbf{S} = \left(f_{S_1}(\mathbf{PA}^{\mathcal{G}}_{S_1}, N_{S_1}), \ldots, f_{S_k}(\mathbf{PA}^{\mathcal{G}}_{S_k}, N_{S_k})\right).$$

Again, one can substitute the parents of $S_i$ by the corresponding functional equations and proceed recursively. After finitely many steps one obtains $\mathbf{S} = f(N_{T_1}, \ldots, N_{T_l})$, where $\{T_1, \ldots, T_l\}$ is the set of *all* ancestors of nodes in $\mathbf{S}$, which does not contain $X$. Since all noise variables are jointly independent we have $N_X \perp\!\!\!\perp \mathbf{S}$. □

**Proof of Lemma 4** According to Definition 4 we can choose $x_\mathbf{S}$, such that $p(x_\mathbf{S}) > 0$ and

$$\left(f_B(x_{\mathbf{PA}^{\mathcal{G}}_B \setminus \{A\}}, \underbrace{\cdot}_{A}, \underbrace{\cdot}_{N_B}), \mathbb{P}^{A \mid X_\mathbf{S} = x_\mathbf{S}}, \mathbb{P}^{N_B}\right) \in \mathcal{A}.$$

Because of $\mathbf{S} \subseteq \mathbf{ND}^{\mathcal{G}}_B$ and Lemma 3 we can apply Lemma 2, which gives $f_B(x_{\mathbf{PA}^{\mathcal{G}}_B \setminus \{A\}}, A|_{X_\mathbf{S}=x_\mathbf{S}}, N_B) = B|_{X_\mathbf{S}=x_\mathbf{S}}$.

But then (1) reads

$$A|_{X_\mathbf{S}=x_\mathbf{S}} \not\perp\!\!\!\perp f_B(x_{\mathbf{PA}^{\mathcal{G}}_B \setminus \{A\}}, A|_{X_\mathbf{S}=x_\mathbf{S}}, N_B) = B|_{X_\mathbf{S}=x_\mathbf{S}}$$
□

**Proposition 2** *If the joint distribution has a strictly positive density with respect to some product measure, Lemma 4 is equivalent to causal minimality.*

**Proof.** Suppose Lemma 4 does not hold. Then

$$\exists \mathbf{S} : \mathbf{PA}^{\mathcal{G}}_B \setminus \{A\} \subseteq \mathbf{S} \subseteq \mathbf{ND}^{\mathcal{G}}_B \text{ and } B \perp\!\!\!\perp A \mid \mathbf{S}$$
$$\Rightarrow \exists \tilde{\mathbf{S}} : B \perp\!\!\!\perp A \mid \mathbf{PA}^{\mathcal{G}}_B \setminus \{A\} \cup \tilde{\mathbf{S}} \text{ and } B \perp\!\!\!\perp \tilde{\mathbf{S}} \mid \mathbf{PA}^{\mathcal{G}}_B$$
$$\stackrel{(*)}{\Rightarrow} \exists \tilde{\mathbf{S}} : B \perp\!\!\!\perp (A, \tilde{\mathbf{S}}) \mid \mathbf{PA}^{\mathcal{G}}_B \setminus \{A\}$$
$$\Rightarrow B \perp\!\!\!\perp A \mid \mathbf{PA}^{\mathcal{G}}_B \setminus \{A\}$$
$$\Rightarrow P(X_\mathbf{V}) = P(B|\mathbf{PA}^{\mathcal{G}}_B \setminus \{A\}) \prod_{X \neq B} P(X|\mathbf{PA}^{\mathcal{G}}_X)$$
$$\Rightarrow P(X_\mathbf{V}) \text{ is Markov wrt to } \mathcal{G} \text{ without } A \to B$$
$$\Rightarrow \text{Causal minimality is violated.}$$
$$\Rightarrow \exists A, B : \text{ is Markov wrt to } \mathcal{G} \text{ without } A \to B$$
$$\Rightarrow \exists A, B : A \perp\!\!\!\perp B \mid \mathbf{PA}^{\mathcal{G}}_B \setminus \{A\}$$
$$\Rightarrow \text{Lemma 4 is violated.}$$

$(*)$ is the "intersection" property of conditional independence [e.g. 1.1.5 in Pearl, 2009] and requires positivity of the densities. □